\documentclass[runningheads]{llncs}

\usepackage{eccv}

\usepackage{eccvabbrv}

\usepackage{graphicx}
\usepackage{booktabs}

\usepackage[accsupp]{axessibility}  

\usepackage[pagebackref,breaklinks,colorlinks,citecolor=eccvblue]{hyperref}

\usepackage{orcidlink}

\usepackage{textcomp}
\usepackage{stfloats}
\usepackage{url}
\usepackage{verbatim}
\usepackage{multirow}
\usepackage{graphicx}
\usepackage{algorithm}
\usepackage{amssymb}
\usepackage{amsmath}
\usepackage{wasysym}
\usepackage{booktabs}

\usepackage{pifont}
\newcommand{\cmark}{\ding{51}}%
\newcommand{\xmark}{\ding{55}}%

\usepackage{color}

\newcommand{\redbf}[1]{{\textbf{\color{red}{#1}}}} 
\newcommand{\blueud}[1]{{\underline{\color{blue}{#1}}}} 

\begin{document}

\title{Anomaly Detection by Adapting a pre-trained Vision Language Model} 

\titlerunning{Anomaly Detection by Adapting a pre-trained Vision Language Model}

\author{Yuxuan Cai\inst{1} \and
Xinwei He\inst{2} \and
Dingkang Liang\inst{1} \and
Ao Tong\inst{1} \and
Xiang Bai\inst{1}}

\authorrunning{Cai.~et al.}

\institute{Huazhong University of Science and Technology \and
Huazhong Agriculture University}

\maketitle

\vspace{-1em}
\begin{abstract}
Recently, large vision and language models have shown their success when adapting them to many downstream tasks. In this paper, we present a unified framework named CLIP-ADA for \textbf{A}nomaly \textbf{D}etection by \textbf{A}dapting a pre-trained CLIP model.
To this end, we make two important improvements: 1) To acquire unified anomaly detection across industrial images of multiple categories, we introduce the learnable prompt and propose to associate it with abnormal patterns through self-supervised learning. 2) To fully exploit the representation power of CLIP, we introduce an anomaly region refinement strategy to refine the localization quality. During testing, the anomalies are localized by directly calculating the similarity between the representation of the learnable prompt and the image.
Comprehensive experiments demonstrate the superiority of our framework, e.g., we achieve the state-of-the-art 97.5/55.6 and 89.3/33.1 on MVTec-AD and VisA for anomaly detection and localization. In addition, the proposed method also achieves encouraging performance with marginal training data, which is more challenging. Code will be available at \href{https://anonymous.4open.science/r/CLIP-ADA/}{https://anonymous.4open.science/r/CLIP-ADA/}.
  \keywords{Anomaly detection \and learnable prompt \and region refinement}
\end{abstract}

\section{Introduction}
\label{sec:intro}

Anomaly detection is a challenging and active research topic in computer vision, boasting widespread industrial applications. The scarcity of abnormal data and the costly data annotation process have steered research towards unsupervised anomaly detection~\cite{bergmann2019mvtec,UniAD}. These approaches focus on identifying anomalous patterns exclusively from the distribution of normal data.

Most research~\cite{MKD,DRAEM,RD4AD} follows a one-model-one-class paradigm which aims to learn a single model for each category separately. Despite impressive results, such a paradigm is resource-intensive. Another promising approach is to learn a model capable of identifying anomalies across multi-class scenarios.
However, the high data diversity makes it hard to learn a joint representation of the normal data. The recent seminal work UniAD~\cite{UniAD} addresses this issue by employing a learnable query to acquire the compact representation of normal data across various categories. DiAD~\cite{he2023diad} employs a diffusion-based~\cite{ho2020denoising} model for image reconstruction, using the reconstruction error as the anomaly detection result.
However, establishing clear anomaly boundaries remains a challenge since the lack of pixel-level supervision. In this paper, we introduce synthetic data to help the model learn discriminative boundaries between normal and abnormal regions.

\begin{figure}[t]
\centering
\includegraphics[width=\linewidth]{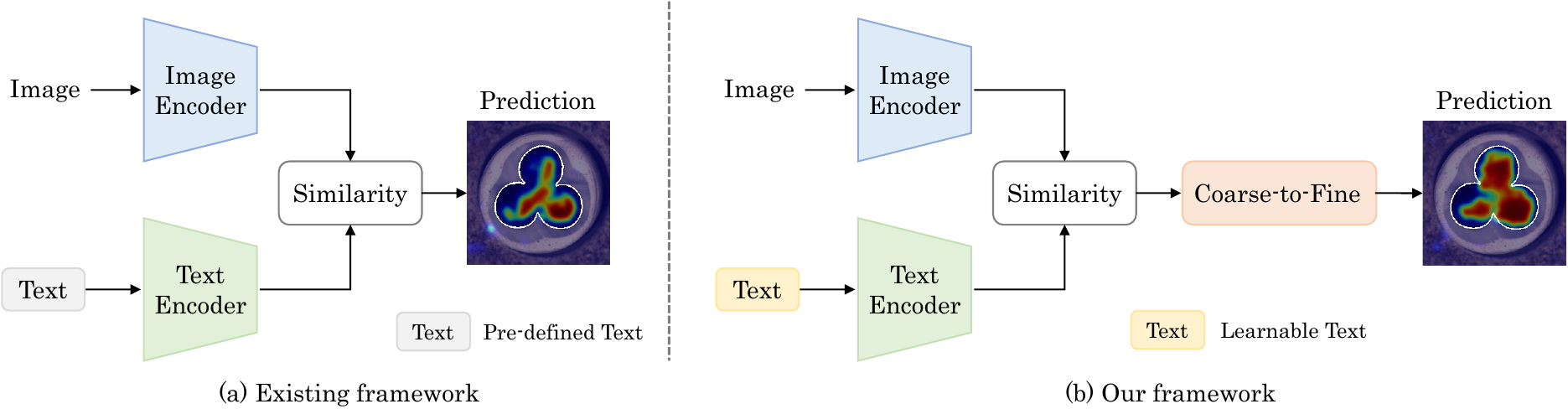}
\caption{(a) Existing CLIP-Based anomaly detection methods, which directly utilize hand-craft text templates and leverage the similarity between the image and text features to spot anomalies; (b) The proposed CLIP-ADA enables more accurate anomaly detection and localization results by incorporating learnable text prompts and a coarse-to-fine strategy.}
\label{fig1}
\end{figure}

Recently, large-scale vision-language (V$\&$L) models like CLIP\cite{CLIP} have been proven their remarkable transferability to downstream tasks\cite{zou2023generalized,zhao2022exploiting, denseclip,xu2022simple} such as object detection and semantic segmentation, benefiting from their capability to extract rich representations. 
More recently, as shown in ~\cref{fig1}, some works~\cite{chen2023zero, jeong2023winclip} attempt to leverage the generalized representations from a pre-trained CLIP for zero-shot/few-shot anomaly detection. Despite encouraging zero-shot/few-shot performance, their performance hardly satisfied the high-accuracy requirements in practical industrial situations. 
Considering the strong representation ability of CLIP, we are motivated to ask: can we maximally adapt the CLIP to a high-performing unified model on unsupervised anomaly detection? 

Inspired by this, we explore how to effectively excavate the high-quality representations of CLIP to the anomaly detection field, which is a challenging task but has not been fully investigated. Firstly, as CLIP is primarily trained on natural images which have a large domain gap with industrial images, directly applying the pre-trained CLIP model may yield undesirable results. Additionally, since CLIP is trained with image-text constraints, it is efficient in capturing discriminative image-level representations. However, the anomaly detection task requires more fine-grained representations to derive precise boundaries of anomaly regions.

To address the aforementioned limitations, we propose a novel framework, termed CLIP-ADA, for multi-class anomaly detection. As shown in \cref{fig1}(b), CLIP-ADA incorporates learnable prompts to acquire a unified representation of abnormal regions, and a Region Refinement module to localize anomalous regions more accurately.
Specifically, since the categories are agnostic in the unsupervised setting, inspired by CoOp\cite{coop}, we employ the learnable prompt to formulate the object-agnostic text prompt. The text prompt is expected to align with the anomalous regions of images. 
Despite these advancements, accurately localizing anomaly regions is still challenging, as anomalies often manifest subtly and bear only negligible differences from normal patterns. To address this, a coarse-to-fine approach is employed. The core idea is utilizing the initial coarse localization result as an attention map, which assists in identifying potential anomalous areas, thereby reducing the focus on irrelevant regions and consequently improving the anomaly detection quality. We summarize our contributions as follows:
\begin{itemize}
    \item We propose a simple yet effective approach to get a unified representation across diverse image categories. Our framework is better suited for varying structures and appearances compared to traditional, manually designed text prompts.
    \item We propose a coarse-to-fine strategy to enhance the accuracy of anomaly localization, by employing the coarse localization result to focus the model on crucial anomalous regions.
    \item Extensive experiments demonstrate the superiority of CLIP-ADA. We outperform existing multi-class anomaly detection methods by 3.0\% and 7.0\% on MVTec~\cite{bergmann2019mvtec} and VisA~\cite{zou2022spot}, respectively.
\end{itemize}

\section{Realted Works}
\subsection{Anomaly Detection}
Most anomaly detection approaches follow a one-model-one-class manner, which can be categorized into three groups:

\textbf{Reconstruction-based methods.} These approaches~\cite{venkataramanan2020attention, liu2022reconstruction, liu2020towards} assume that the model trained solely on normal data will struggle to accurately reconstruct anomalous regions. As a result, the anomaly region can be spotted according to the reconstruction error. However, this hypothesis may be invalid due to the ``identical shortcut'' problem~\cite{UniAD}, where the reconstruction result will always equal the input image. 

\textbf{Embedding-based methods.} Commonly, these approaches~\cite{wan2021industrial,reiss2021panda,padim,patchcore} employ pre-trained models to extract representations of normal images, subsequently compressing these representations into a specific space through clustering techniques. During the testing phase, anomalies are recognized based on the distance from the test image to the established space. However, these methods face mismatch problems since there exists a domain gap between the pre-trained dataset and the anomaly dataset. Moreover, the necessity of the memory bank to store the normal feature representations is memory-unfriendly and limits the inference speed.

\textbf{Synthesizing-based methods.} These methods~\cite{schluter2022natural,DRAEM,BiaS,zavrtanik2022dsr,li2021cutpaste} generate anomalous samples during training to guide the network to effectively distinguish between normal and anomalous data, thereby establishing distinct decision boundaries. For instance, DRÆM~\cite{DRAEM} and NSA~\cite{schluter2022natural} propose customized strategies to generate close-to-real synthetic samples, and subsequently train localization networks using segmentation procedures. DSR~\cite{zavrtanik2022dsr} generates anomalous data at the feature level. However, due to the diversity of anomalies, it is challenging to comprehensively cover all anomaly types with synthetic anomalies.

\subsection{Unified Anomaly Detection}
UniAD~\cite{UniAD} first proposes the unified anomaly detection task, which aims to train a single model capable of spotting anomalies in diverse industrial images. UniAD addresses the more severe ``identical shortcut'' problem of reconstruction networks in the unified setting. They claim that the query embedding in the attention layer is useful to prevent the model from learning shortcuts. Therefore, a transformer-based reconstruction network is proposed, incorporating a layer-wise query decoder, to extensively apply query embedding. OmniAL~\cite{zhao2023omnial} is also based on the idea of reconstruction, which solves the ``identical shortcut'' from input data. It proposes an effective panel-guided synthesis strategy to perturb normal data, and train a network to reconstruct normal. Most recently, DiAD~\cite{he2023diad} proposes a diffusion-based framework for reconstruction, and introduces a latent-space Semantic-Guided network to preserve image’s category information.

\subsection{Vision-Language Contrastive Learning}

The vision-language pre-training model, CLIP~\cite{CLIP}, aims to build the connection between the vision and language modalities. Many CLIP-powered methods~\cite{proposaclip,denseclip,xu2022simple} have been proposed on various downstream tasks, such as object detection~\cite{proposaclip} and semantic segmentation~\cite{denseclip}.
Some recent works~\cite{jeong2023winclip, chen2023zero, zhou2023anomalyclip} turn CLIP to zero-shot anomaly detection. 
For instance, WinCLIP~\cite{jeong2023winclip} directly leverages a frozen pre-trained CLIP model to spot anomalies. They first design hand-crafted text templates for the anomaly detection task. Then propose a window-based strategy to extract and aggregate image features. Finally localizing anomalies through the similarity between text embedding and image features. Based on WinCLIP, VAND~\cite{chen2023zero} introduces linear projection for fine-tuning to make features adaptable to anomaly detection scenarios, and thus derive more accurate localization results. However, the zero-shot approach demonstrates a relatively weaker performance. In this paper, we present a simple yet effective framework to adapt the CLIP for anomaly detection.

\section{Preliminary}
In this section, we revisit CLIP and then give the problem definition of unified anomaly detection. 

\subsection{A revisit of CLIP}
CLIP, a prototypical pre-trained vision-language model, contains an image encoder for encoding image representations and a text encoder for transforming text inputs into text embeddings. During its pre-training phase, CLIP collects 400 million image-text pairs, and then aligns these paired image and text features through self-supervised learning learning.

The pre-trained CLIP can be easily transferred to the downstream classification task. For example, we can define a set of text prompts such as ``\texttt{a photo of a [CLS]}.'', where \texttt{[CLS]} is an actual class name. Then, for a given image, CLIP will compute the semantic similarity between the encoded image representation and text embeddings, and the prediction corresponds to the class with the highest score. In this paper, we aim to extend the alignment property of CLIP to the unified anomaly detection task~\cite{UniAD}.

\subsection{Problem Definition}
Anomaly detection techniques focus on distinguishing images as either normal or abnormal, and accurately localizing abnormal areas.
Given a training dataset $D_{train} = \{D_1, D_2, \cdots, D_K\}$ and a test dataset $D_{test} = \{{D'}_1, {D'}_2, \cdots, {D'}_K\}$, where $K$ denotes the category number. 
The traditional anomaly detection task follows a one-model-one-class setting, where a model is trained for each specific dataset $D_k$($k \in \{1, \cdots, K\}$) and tested on its corresponding test set ${D'}_k$. 
In contrast, the unified anomaly detection task~\cite{UniAD} utilizes a one-model-multi-class scheme, where a single model is trained on the entire training dataset $D_{train}$. This single model is expected to handle all categories simultaneously during inference. In this paper, we focus on the unified anomaly detection task. 

\section{Method}
\subsection{Frozen CLIP as the Backbone}\label{sec_backbone}

We utilize a pre-trained CLIP~\cite{CLIP} as the backbone of our framework to extract representations for downstream anomaly detection. The process is briefly introduced as follows.

For the image encoder, CLIP supports both ViT~\cite{vit} and ResNet~\cite{resnet}. In practice, we use the pre-trained ViT to extract image features. Given an input image $I \in \mathbb{R}^{H \times W \times 3}$, where $H,W$ denotes the height and width, the CLIP model first reshapes the image to a sequence of 2D patches $I_p \in \mathbb{R}^{N_p \times S^2 \times 3}$, where $S$ and $N_p$ represent the patch size and the number of patches, respectively. Then the transformer layer will project $I_p$ to the embedding feature, we denote the feature from the $i$-th stage as $\mathcal{F}_{i} \in \mathbb{R}^{N_p \times D}$, where $D$ denotes the embedding dimension.  We believe that both low-level and high-level representations are vital for anomaly detection, since some anomalies are distinct from normal patterns in appearance, while some are required for contextual information to detect. Thus, we use a specific layer ($i=7$) to produce the visual representation $\mathcal{F}_{clip}$, which takes both detail and semantic information into consideration. 
In addition, we append a linear projection layer $\Psi_{0}$ after the image encoder to fine-tune $\mathcal{F}_{clip}$ for better adaptation to anomaly detection task, denoted by $\mathcal{F} = \Psi_{0}(\mathcal{F}_{clip}) \in \mathbb{R}^{N_p \times C}$, where $C$ is the projection embedding dimension.

For the text encoder, we input the text template and obtain continuous vectors $\mathcal{V}$ as its outputs. Generally, the text template is composed of pre-defined text templates and names of objects, such as ``{\tt A photo of a [CLS]}''. Inspired by WinCLIP~\cite{jeong2023winclip}, we incorporate a task-specific description to the pre-defined template to provide more contextual information in the text input. Specifically, a straightforward text template is defined as ``{\tt A photo of a [CLS] with defects for anomaly detection}''. 

However, we argue that it is hard to accurately define anomalies by the hand-craft text templates, and limited fixed templates are difficult to cope with anomalies of different shapes and appearances. 
As a result, we need a strategy that can adaptively learn anomaly representations.

\begin{figure*}[t]
\centering
\includegraphics[width=\textwidth]{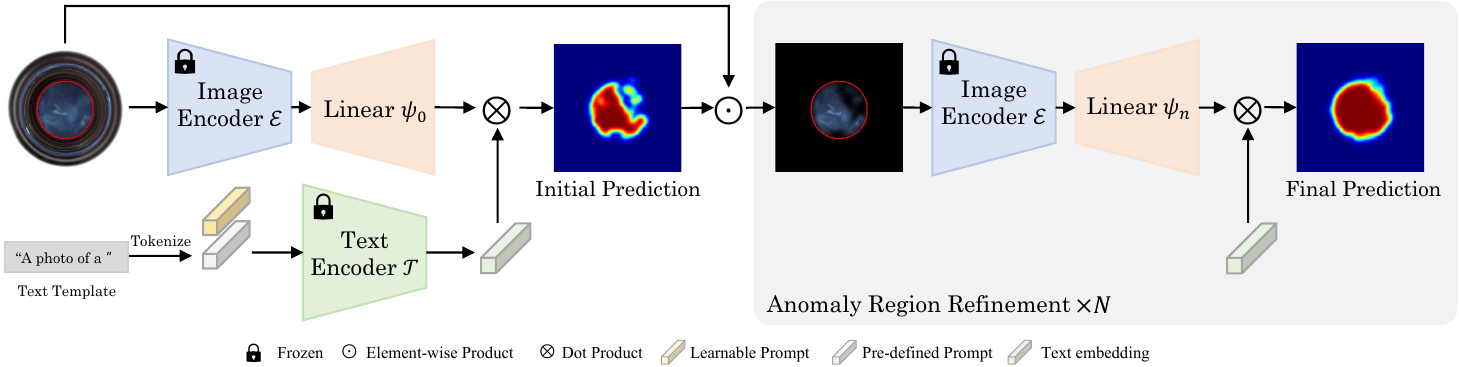}
\caption{\textcolor{black}{An overview of our method. We use the frozen CLIP as our backbone. In the image branch, we extract visual features through the CLIP image encoder, followed by a linear layer for adaptation. In the text branch, we combine the pre-defined and learnable prompts, and then employ a CLIP text encoder to derive the text embedding. Subsequently, we calculate the similarity map between the text embedding and visual features, which acts as the initial localization result. To improve the quality of the initial prediction, we further propose a refinement strategy.}}
\label{overviewpipe}
\end{figure*}

\subsection{Anomaly Adaptation with Learnable Prompts}\label{sec:learnable_prompt}

In order to acquire a unified and robust representation for all categories, inspired by CoOp~\cite{coop}, we remove the original {\tt [CLS]} from the template and incorporate learnable prompts into it. The process can be formulated as:

\begin{equation}
    t' = Emb(Tokenize(Text)) \in \mathbb{R}^{K \times L \times D},
\end{equation}
where $Text$ denotes the pre-defined text template. $Tokenize$ converts the template to unique IDs, and $Emb(\cdot)$ denotes the linear embedding layer, which is used to project IDs into embedding space. $K$ denotes the number of categories, which is set to 1 under the unified setting. $L$ is the length of the template, $D$ is set to 512 empirically. After obtaining the word embedding $t'$ of the text template, we then insert the learnable prompts into it to form the input of the text encoder $\mathcal{T}$, which is formulated as:
\begin{equation}
    t = [t'_0, \cdots, t'_x, P_0, P_1, \cdots, P_{S-1}, t'_{x+1}, \cdots, t'_{L-1}],
\end{equation}
where $x$ represents that we insert the learnable prompts at location $x$, each $P_s (s \in \{0, \cdots, S-1\})$ is a vector with dimension $D$, \emph{i.e.}, $P_s \in \mathbb{R}^{K \times D}$. $S$ denotes the length of the learnable prompts, which is set to 4 in practice. Then the output of the text encoder can be calculated by $\mathcal{V} = \mathcal{T}(t) \in \mathbb{R}^{K \times C}$, where $C$ is set to 1024 in CLIP.
 
\subsubsection{Optimizing Objective.} 
In order to drive the learnable prompts to associate with the anomaly regions, we propose to adopt self-supervised learning. Specifically, we first perturb the given normal image $I$ to obtain the synthetic anomaly image, along with the corresponding pixel-level ground-truth. Following this, we calculate the pixel-level similarity map between the text embedding $\mathcal{V}$ and the visual representation $\mathcal{F}$, which is defined as:

\begin{equation}
\label{coarsemap}
    M = Sigmoid( \mathcal{F} \cdot \mathcal{V}^{T}) \in \mathbb{R}^{K \times N_p},
\end{equation}
we use a sigmoid activation to normalize the alignment result and then reshape $M$ to $M \in \mathbb{R}^{s_p \times s_p}$, where $s_p = \sqrt{N_p}$. We aim to minimize the binary cross-entropy (BCE) loss between the similarity map $M$ and its corresponding ground-truth, which is formulated as:
\begin{equation}
\label{alignloss}
    \mathcal{L}_{align} = \sum y_i log(x_i) + (1 - y_i)log(1 - x_i),
\end{equation}
where $x_i$ and $y_i$ are the predicted similarity and the corresponding label, respectively.

\subsection{Anomaly Region Refinement}\label{sec:anomaly_refine}

Although learnable prompts can help spot the anomaly regions with high accuracy, it still faces great challenges when the anomaly regions are subtle or negligible against the normal regions. 
We assume this is because the learnable prompts are trained to match the anomaly regions globally and are prone to produce erroneous results at the local boundary areas. To address this issue, we introduce a simple coarse-to-fine strategy to guide the model to pay more attention to the coarse prediction regions, thereby facilitating the learning of discriminative representations to delineate the boundary more accurately.

Specifically, we leverage the above initial output $M$ as an attention map and perform element-wise product with the input image $I$ by down-sampling $I$ to the size of $M$, \emph{i.e.}, $I_{e} = M \odot I$. Recall that $M$ is trained to have high responses at anomaly regions while low responses at normal regions. Thus this operation helps the network to discover more fine-grained details near the anomaly regions to better delineate them. 
Subsequently, the enhanced image $I_{e}$ is forwarded to the CLIP image encoder and the projection layer to acquire the enhanced visual representation $\mathcal{F}_e \in \mathbb{R}^{N_p \times C}$. 

Finally, we adopt the BCE loss to align the enhanced representation $\mathcal{F}_{e}$ with the text embedding $\mathcal{V}$ in anomalous regions. Concretely, the similarity between $\mathcal{F}_{e}$ and $\mathcal{V}$ is defined as:
\begin{equation}
\label{refinemap}
    M_e = Sigmoid(\mathcal{F}_e \cdot \mathcal{V}^{T}) \in \mathbb{R}^{K \times N_p},
\end{equation}
The training objective $\mathcal{L}^e_{align}$ can be defined in the same way as Equation (\ref{alignloss}). Please note that we can adopt the refinement layer $N$ times, and the linear projection layers $\Psi_i (i=0,1,\cdots, N-1)$ do not share weights.
\subsection{Overall Objective and Inference}

\subsubsection{Overall Objective.} Our whole framework is trained end-to-end. During training, we combine the above loss functions as follows: 
\begin{equation}
    \mathcal{L} = \mathcal{L}_{align} + \lambda \mathcal{L}^{e}_{align},
\end{equation}
 where $\lambda$ is a trade-off hyper-parameters to balance the loss between the initial prediction and the refinement prediction. We set $\lambda$ to 1 in this paper.
\subsubsection{Inference.} For each image $I \in D_{test}$, we first obtain the localization result based on the image features and text prompt as described in Equation~(\ref{refinemap}).
Subsequently, we interpolate it to match the spatial resolution of the input $I$ and apply a Gaussian filter for smoothing, yielding the pixel-level anomaly score map $M_s$. Additionally, We acquire the image-level anomaly score by calculating the average of the top $K$ scores in the $M_s$. $K$ is set to 500 experimentally.

\section{Experiments}
\subsection{Experimental Setups}

\subsubsection{Datasets}  We conduct extensive experiments on two anomaly detection datasets, \emph{i.e.},  MVTec-AD (MVTec)~\cite{bergmann2019mvtec} and VisA~\cite{zou2022spot}. MVTec is widely recognized and used, while VisA is notably more challenging than MVTec.

MVTec consists of 15 different categories. The training set contains 3,629 normal images, while the test set includes 1,725 images covering both normal and anomaly data. The anomalous regions are annotated at the pixel level.

VisA consists of 12 different categories. It contains 8,659 normal images for training, 2,162 images covering 962 normal images and 1,200 anomalous images for testing. Pixel-level annotations are provided. 
\subsubsection{Metrics} Following the previous common practice,  we use pixel-level/image-level Area Under the Receiver Operating Curve (P-AUC/I-AUC) for model evaluation. We also adopt pixel-level mean Average Precision (P-mAP) to better evaluate model's localization capability.

\begin{table}[t]
\centering
\caption{Detection performance comparison on MVTecAD. $^\ast$ denotes the reproduced results using the official code. \redbf{Red} and \blueud{blue} indicate the best and the second best, respectively. I-AUC is reported.}
\label{mvtec_results_img}
\resizebox{\textwidth}{!}{
\begin{tabular}{lccccccccccccccc|c}
\toprule
Method & Bottle & Cable & Capsule & Hazel. & Metal. & Pill & Screw & Tooth. & Trans. & Zipper & Carpet & Grid & Leather & Tile & Wood & Mean \\
\midrule
DRÆM                & 97.5   & 57.8  & 65.3    & 93.7     & 72.8       & 82.2 & 92.0    & 90.6       & 74.8       & 98.8   & 98.0     & \redbf{99.3} & 98.7    & \blueud{99.8} & \redbf{99.8} & 88.1 \\
RD4AD                & 99.6   & 84.1  & \blueud{94.1}    & 60.8     & \redbf{100.}        & \redbf{97.5} & \blueud{97.7}  & 97.2       & \blueud{94.2}       & \blueud{99.5}   & 98.5   & 98.0   & \redbf{100.}     & 98.3 & 99.2 & 94.6 \\
UniAD$^\ast$                & \blueud{99.7}   & \redbf{95.2}  & 86.9    & \redbf{99.8}     & \blueud{99.2}       & 93.7 & 87.5  & 94.2       & \redbf{99.8}       & 95.8   & \blueud{99.8}   & 98.2 & \redbf{100.}     & 99.3 & 98.6 & 96.5 \\
DenoiseAD            & 97.9   & 89.0    & 73.6    & 92.3     & \blueud{99.2}       & 64.7 & 89.9  & 87.5       & 92.5       & 85.5   & \redbf{100.}    & \redbf{99.3} & \redbf{100.}     & 98.0   & 97.9 & 91.2 \\
DiAD                 & \blueud{99.7}   & \blueud{94.8}  & 89.0      & \blueud{99.5}     & 99.1       & 95.7 & 90.7  & \redbf{99.7}       & \redbf{99.8}       & 95.1   & 99.4   & 98.5 & 99.8    & 96.8 & \blueud{99.7} & \blueud{97.2} \\
Ours                 & \redbf{99.8}   & 94.0    & \redbf{94.3}    & 99.2     & 98.1       & \blueud{95.9} & \redbf{99.4}  & \blueud{98.9}       & 92.3       & \redbf{100.}    & 98.6   & \blueud{98.8} & \blueud{99.9}    & \redbf{100.}  & 93.5 & \redbf{97.5} \\
\bottomrule
\end{tabular}}
\end{table}

\subsubsection{Implementation Details}  We resize all the images to $224 \times 224$. The image encoder is instantiated with the
pre-trained ViT-B/16 of the CLIP. The pre-trained CLIP model is frozen, while the remaining components are trained using the AdamW optimizer. For MVTec, the model is trained for 800 epochs, and the learning rate is set to $2e-4$ and multiplied by 0.2 at 400-$th$ and 700-$th$ epochs. The batch size is set to 16. For VisA, the epoch is set to 500, the learning rate is set to $4e-4$ and multiplied by 0.2 at 250-$th$ epoch. The batch size is set to 64.
For both MVTec and VisA, we employ the synthesis strategy proposed in DRÆM~\cite{DRAEM}. Please refer to Supplementary for more details. 

\vspace{-0.5em}
\subsubsection{Comparison Methods} 
For comprehensive comparisons, we compare our framework with the following representative anomaly detection methods: 
1)  DRÆM~\cite{DRAEM} and RD4AD~\cite{RD4AD}, which are designed for the One-Model-One-Class paradigm, 2) UniAD~\cite{UniAD}, DenoiseAD and DiAD~\cite{he2023diad}, which follow the One-Model-Multi-Class paradigm. In this paper, all methods are trained under the One-Model-Multi-Class setting.

\begin{table}[t]
\centering
\caption{Localization performance comparison on MVTecAD. $^\ast$ denotes the reproduced results using the official code. \redbf{Red} and \blueud{blue} indicate the best and the second best, respectively.}
\label{mvtec_results}
\resizebox{\textwidth}{!}{
\begin{tabular}{c|c|c|c|c|c|c}
\toprule
\multirow{2}{*}{Category} & \begin{tabular}[c]{@{}c@{}}DRÆM\\      (ICCV 2022)\end{tabular} & \begin{tabular}[c]{@{}c@{}}RD4AD\\      (CVPR 2022)\end{tabular} & \begin{tabular}[c]{@{}c@{}}UniAD$^\ast$\\      (NeurIPS 2022)\end{tabular} & \begin{tabular}[c]{@{}c@{}}DenoiseAD\\      (ICCV 2023)\end{tabular} & \begin{tabular}[c]{@{}c@{}}DiAD\\      (AAAI 2024)\end{tabular} & Ours        \\\cmidrule{2-7}
                          & \multicolumn{6}{c}{P-AUC / P-mAP}                                           \\
\midrule
Bottle                    & 87.6 / 62.5                                                      & 97.8 / \blueud{68.2}                                                      & \blueud{98.1} / 66.0                                                           & 97.7 /  -                                                            & \redbf{98.4} / 52.2                                                     & 96.6 / \redbf{69.7} \\
Cable                     & 71.3 / 14.7                                                      & 85.1 / 26.3                                                      & \redbf{97.3} / 39.9                                                         & 95.2 /  -                                                            & \blueud{96.8} / \blueud{50.1}                                                     & 95.3 / \redbf{59.4} \\
Capsule                   & 50.5 / 6.0                                                         & \redbf{98.8} / \redbf{43.4}                                                      & \blueud{98.5} / \blueud{42.7}                                                         & 98.0 /  -                                                              & 97.1 / 42.0                                                       & 97.2 / 30.2 \\
Hazelnut                  & 96.9 / \blueud{70.0}                                                        & 97.9 / 36.2                                                      & \blueud{98.1} / 55.2                                                         & 97.7 /  -                                                            & \redbf{98.3} / \redbf{79.2}                                                     & 97.9 / 52.9 \\
Metal\_nut                & 62.2 / 31.1                                                      & 93.8 / \redbf{62.3}                                                      & 94.8 / 55.5                                                         & \blueud{96.8} /  -                                                            & \redbf{97.3} / 30.0                                                       & 93.8 / \blueud{59.5} \\
Pill                      & 94.4 / 59.1                                                      & \blueud{97.5} / \blueud{63.4}                                                      & 95.0 / 44.0                                                             & 92.5 /  -                                                            & 95.7 / 46.0                                                       & \redbf{97.8} / \redbf{65.5} \\
Screw                     & 95.5 / 33.8                                                      & \redbf{99.4} / \blueud{40.2}                                                      & 98.3 / 28.7                                                         & \blueud{99.0} /  -                                                              & 97.9 / \redbf{60.6}                                                     & 93.5 / 9.2  \\
Toothbrush                & 97.7 / 55.2                                                      & \redbf{99.0} / 53.6                                                        & 98.4 / 34.9                                                         & \blueud{98.9} /  -                                                            & \redbf{99.0} / \redbf{78.7}                                                       & \blueud{98.9} / \blueud{69.4} \\
Transistor                & 64.5 / 23.6                                                      & 85.9 / \blueud{42.3}                                                      & \redbf{97.9} / \redbf{59.5}                                                         & 92.6 /  -                                                            & \blueud{95.1} / 15.6                                                     & 86.5 / 38.0   \\
Zipper                    & 98.3 / \redbf{74.3}                                                      & \blueud{98.5} / 53.9                                                      & 96.8 / 40.1                                                         & 97.6 /  -                                                            & 96.2 / 60.7                                                     & \redbf{98.9} / \blueud{66.4} \\
Carpet                    & 98.6 / \blueud{78.7}                                                      & \blueud{99.0} / 58.5                                                        & 98.5 / 49.9                                                         & 98.9 /  -                                                            & 98.6 / 42.2                                                     & \redbf{99.5} / \redbf{80.8} \\
Grid                      & 98.7 / 44.5                                                      & \redbf{99.2} / \blueud{46.0}                                                        & 96.5 / 23.0                                                           & \blueud{99.1} /  -                                                            & 96.6 / \redbf{66.0}                                                       & 98.0 / 23.3   \\
Leather                   & 97.3 / \redbf{60.3}                                                      & \blueud{99.3} / 38.0                                                        & 98.8 / 32.9                                                         & \redbf{99.5} /  -                                                            & 98.8 / \blueud{56.1}                                                     & \blueud{99.3} / 47.1 \\
Tile                      & \blueud{98.0} / \redbf{93.6}                                                        & 95.3 / 48.5                                                      & 91.8 / 42.1                                                         & 92.1 /  -                                                            & 92.4 / 65.7                                                     & \redbf{99.0} / \blueud{90.9}   \\
Wood                      & \blueud{96.0} / \redbf{81.4}                                                        & 95.3 / 47.8                                                     & 93.2 / 37.2                                                         & 94.7 /  -                                                            & 93.3 / 43.3                                                     & \redbf{97.7} / \blueud{72.4} \\\hline
Mean                      & 87.2 / 52.5                                                      & 96.1 / 48.6                                                      & \redbf{96.8} / 43.4                                                         & \blueud{96.7} /  -                                                            & \redbf{96.8} / \blueud{52.6}                                                     & \blueud{96.7} / \redbf{55.6} \\
\bottomrule
\end{tabular}}
\end{table}

\begin{figure*}[]
\centering
\includegraphics[width=0.95\textwidth]{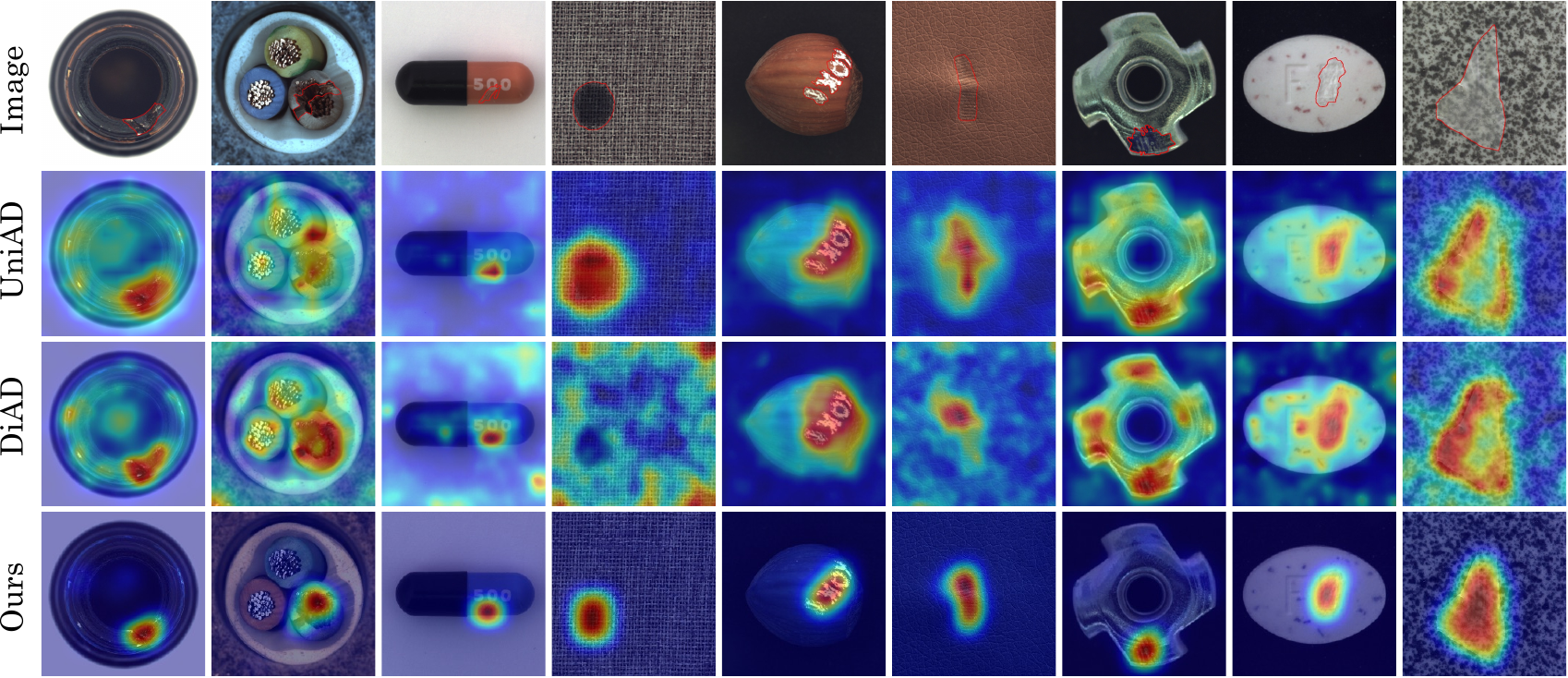}
\caption{\textcolor{black}{Visualization of localization results on MVTec.}}
\label{visual}
\end{figure*}

\subsection{Comparision with State-of-the-Art}
\subsubsection{Evaluation on MVTec} 

\cref{mvtec_results_img} reports the anomaly detection results on MVTec. It can be observed that we achieve the best performance of 97.5\% I-AUC, outperforming the previous art DiAD~\cite{he2023diad} and UniAD~\cite{UniAD} by 0.3\% and 1.0\%, respectively. Additionally, DRÆM~\cite{DRAEM}, which also employs the synthesizing-based framework, reaches only 88.1\%. In contrast, our method significantly exceeds it by 9.4\%. We attribute this gain to the generalizable features extracted by CLIP, ensuring our framework performs well under the multi-class scenario.

\cref{mvtec_results} compares the anomaly localization performance. We can observe that in terms of localization, UniAD and DiAD are two competitive methods, achieving 96.8\% P-AUC, while ours is slightly lower than theirs by 0.1\%. However, we significantly outperform UniAD and DiAD by 12.2\% and 3.0\% in P-mAP, respectively. \cref{visual} displays qualitative results of UniAD, DiAD, and our method, demonstrating that we achieve the most accurate localization across various anomalies.
Furthermore, we note that in the ``screw'' category, the P-mAP is relatively low, reflecting undesirable localization ability. We hypothesize that the reason is that the location of ``screw'' object is constantly changing, which may prevent the text prompts from covering all normal patterns. Thus, the bridge between the text and image features may not accurately determine normal and abnormal.
Please refer to Supplementary for more details.

\subsubsection{Evaluation on VisA}
\begin{figure*}[t]
\centering
\includegraphics[width=0.95\textwidth]{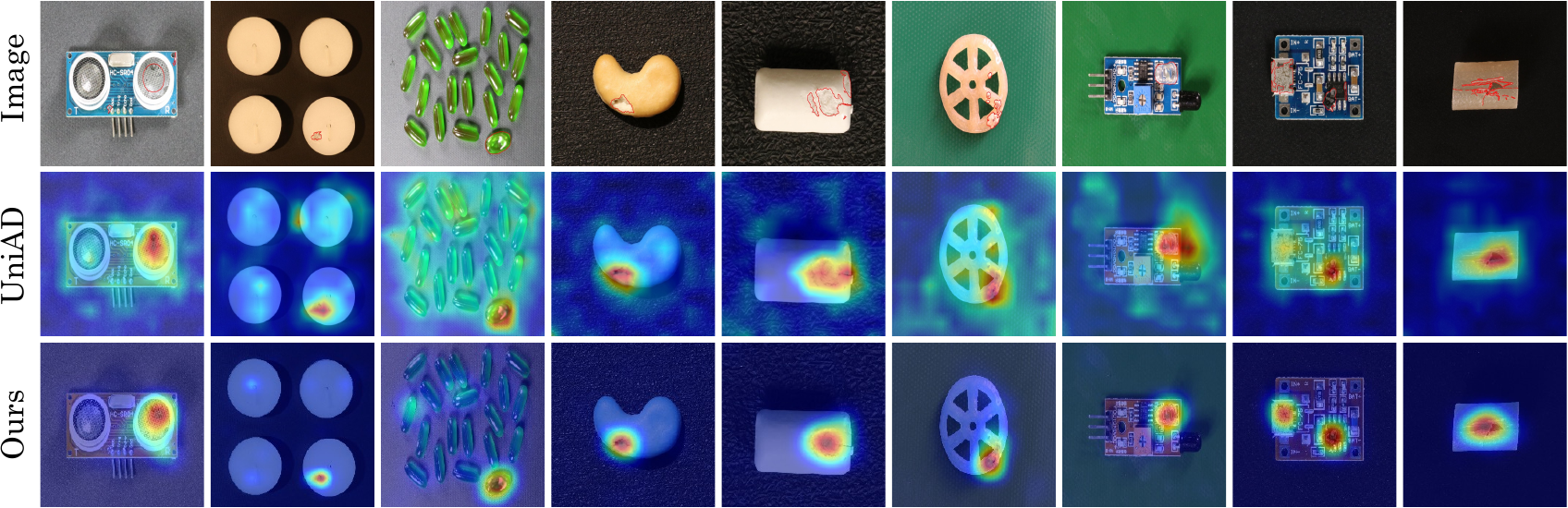}
\caption{\textcolor{black}{Visualization of localization results on VisA.}}
\label{visa_visual}
\end{figure*}

We further compare our method with previous superior works on the challenging VisA dataset. The results are summarized in \cref{visa_results}.  As shown, among the compared methods, we achieve the best results in both anomaly detection and localization, surpassing the previous state-of-the-art DiAD~\cite{he2023diad} by 2.5\% I-AUC, 0.3\% P-AUC and 7.0\% P-mAP, respectively. Compared to UniAD, we can also observe that we comprehensively outperform it.
~\cref{visa_visual} visualizes localization results of our framework and UniAD. It once again proves that our method identifies anomaly regions more faithfully than compared methods.
However, we also note that in categories like ``Macaroni1'', ``Macaroni2'', and ``Candles'', our P-mAP scores are relatively low. We believe the reason is that the text prompt might struggle to learn the fine-grained details required for localization in changing scenes.
Please refer to Supplementary for more details.

\begin{table}[t]
\centering
\caption{Detection and localization performance comparison on VisA. $^\dagger$ denotes results are taken from ~\cite{he2023diad}. \redbf{Red} and \blueud{blue} indicate the best and the second best, respectively.}
\label{visa_results}
\setlength{\tabcolsep}{8pt}
\resizebox{0.95\textwidth}{!}{
\begin{tabular}{c|c|c|c|c}
\toprule
\multirow{2}{*}{Category} & DRÆM$^\dagger$              & UniAD$^\dagger$              & DiAD$^\dagger$             & Ours               \\\cmidrule{2-5}
                          & \multicolumn{4}{c}{I-AUC / P-AUC /  P-mAP}                                     \\
\midrule
PCB1                      & 71.9 / 94.6 / 31.8 & \redbf{92.8} / 93.3 / 3.9  & 88.1 / \blueud{98.7} / \blueud{49.6} & \blueud{92.5} / \redbf{98.8} / \redbf{51.0} \\
PCB2                      & 78.4 / 92.3 / 10.0 & \blueud{87.8} / 93.9 / 4.2  & \redbf{91.4} / \blueud{95.2} / \blueud{7.5} & 82.4 / \redbf{96.9} / \redbf{17.4} \\
PCB3                      & 76.6 / 90.8 / \blueud{14.1} & 78.6 / \redbf{97.3} / 13.8 & \blueud{86.2} / \blueud{96.7} / 8.0 & \redbf{86.7} / 93.8 / \redbf{28.6} \\
PCB4                      & 97.3 / 94.4 / \redbf{31.0} & \blueud{98.8} / 94.9 / 14.7 & \redbf{99.6} / \redbf{97.0} / 17.6 & 95.2 / \blueud{96.8} / \blueud{26.5} \\
Macaroni1                 & 69.8 / 95.0 / \redbf{19.1} & 79.9 / \blueud{97.4} / 3.7  & \blueud{85.7} / 94.1 / \blueud{10.2} & \redbf{95.3} / \redbf{98.0} / 5.4  \\
Macaroni2                 & 59.4 / 94.6 / \redbf{3.9}  & \blueud{71.6} / \blueud{95.2} / 0.9  & 62.5 / 93.6 / 0.9 & \redbf{79.4} / \redbf{95.3} / \blueud{1.9}  \\
Capsules                  & \redbf{83.4} / 97.1 / \blueud{27.8} & 55.6 / 88.7 / 3.0  & 58.2 / \blueud{97.3} / 10.0 & \blueud{78.8} / \redbf{98.4} / \redbf{38.0} \\
Candles                   & 69.3 / 82.2 / 10.1 & \blueud{94.1} / \redbf{98.5} / \redbf{17.6} & 92.8 / \blueud{97.3} / \blueud{12.8} & \redbf{94.7} / 91.5 / 6.4  \\
Cashew                    & 81.7 / 80.7 / 9.9  & \redbf{92.8} / \redbf{98.6} / 51.7 & \blueud{91.5} / 90.9 / \blueud{53.1} & 88.9 / \blueud{94.0} / \redbf{55.9} \\
Chewing gum               & 93.7 / 91.0 / \blueud{62.3} & \blueud{96.3} / \redbf{98.8} / 54.9 & \redbf{99.1} / 94.7 / 11.9 & 93.6 / \blueud{98.4} / \redbf{62.8} \\
Fryum                     & 89.1 / 92.4 / 38.8 & 83.0 / \blueud{95.9} / 34.0 & \blueud{89.8} / \redbf{97.6} / \redbf{58.6} & \redbf{91.9} / \blueud{95.9} / \blueud{41.5} \\
Pipe fryum                & 82.8 / 91.1 / 38.1 & \blueud{94.7} / \blueud{98.9} / 50.2 & \redbf{96.2} / \redbf{99.4} / \redbf{72.7} & 91.8 / 98.4 / \blueud{61.3} \\\hline
Mean                      & 79.1 / 91.3 / 23.5 & 85.5 / 95.9 / 21.0 & \blueud{86.8} / \blueud{96.0} / \blueud{26.1} & \redbf{89.3} / \redbf{96.3} / \redbf{33.1} \\
\bottomrule
\end{tabular}}
\end{table}

\subsection{Ablation Studies and Analysis}

In this section, we will perform discussions on the essential components of the framework, including anomaly region refinement layers and learnable prompts. The results are summarized in \cref{tablearnrefine}. Furthermore, we will also analyze the performance when fine-tuning the CLIP encoder, with the performance given in \cref{tabtuneclip}. Finally, we explore the influence of using different data scales for training, as shown in \cref{tabdatascale}.

\subsubsection{Anomaly Region Refinement Layers}
\begin{figure}[t]
\centering
\includegraphics[width=0.98\linewidth]{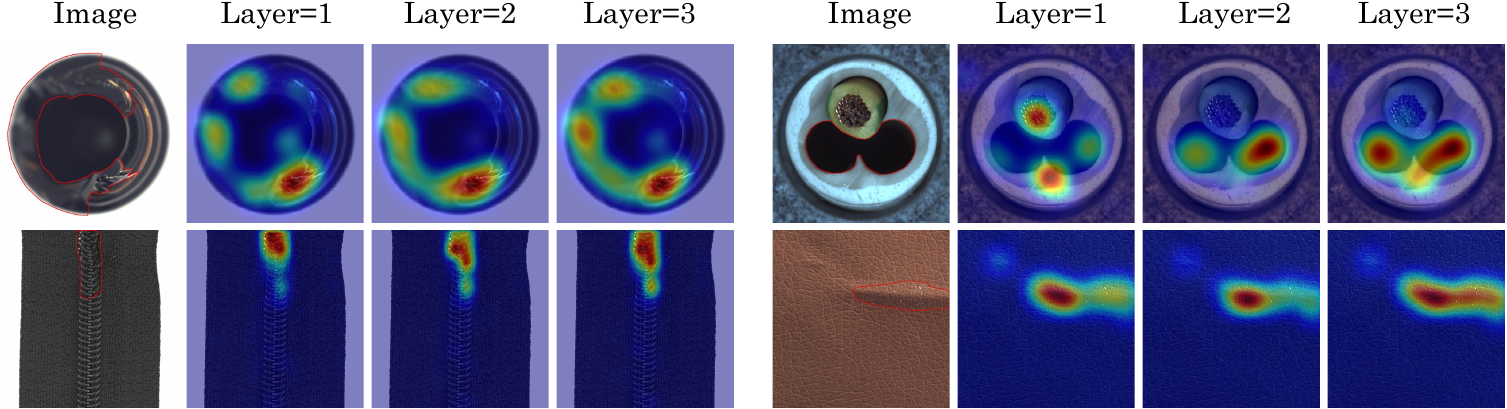}
\caption{Visualization of localization results with different refinement layers.}
\label{figrefine}
\end{figure}

To investigate the importance of Anomaly Region Refinement Layers, we conduct experiments by gradually increasing the Anomaly Region Refinement layers. As shown in \cref{tablearnrefine}, without the anomaly region refinement layer, the framework can obtain P-mAP of 54.2\% and 54.3\% when exploiting fixed and learnable prompts, respectively. While adopting one refinement layer leads to gains of 0.6\% and 1.3\% in P-mAP, respectively. Further increasing the layer to two, the performance slightly increases by 0.2\%/0.1\%. It demonstrates the effectiveness of the refinement strategy in refining the coarse maps and producing more accurate anomaly localization results. 
In terms of P-AUC, we can observe that in the absence of the learnable prompt, performance gradually increases from 96.2\% to 96.6\% as the number of refinement layers increases. 
Meanwhile, it is worth noting that, for both P-AUC and P-mAP, the improvement from zero to one layer is more significant than from one to two layers. We believe that with just one refinement layer, the framework has already learned to localize anomalous regions quite decently, therefore subsequent additions of refinement layers are likely to yield diminishing gains. 
As shown in ~\cref{figrefine}, we can also observe that the model achieves more accurate localization with one refinement layer compared with the baseline without the refinement layer, while adding an extra layer proves beneficial for further enhancing the localization quality. 

It should also be mentioned that the computational complexity also rises progressively. Therefore, considering both the performance and computational complexity of the model, we default use learnable prompts with only one refinement layer for a better trade-off. More details are available in the Supplementary.

\begin{table}[t]
\centering
\caption{Localization performance comparison under different refinement layers and types of prompts. \emph{Num} denotes the number of refinement layers.}
\label{tablearnrefine}
\resizebox{\textwidth}{!}{
\begin{tabular}{ccccccccccccccccccc}
\toprule
\multirow{2}{*}{} & \multirow{2}{*}{$Num$} & \multirow{2}{*}{Learn} & \multicolumn{15}{c}{Category}                                                                                                            & \multirow{2}{*}{Mean} \\
                         &                         &                            & Bot. & Cab. & Cap. & Haz. & Met. & Pil. & Scr. & Too. & Tran. & Zip. & Carp. & Gri. & Lea. & Til. & Woo. &                       \\
\midrule
\multirow{6}{*}{\rotatebox{90}{P-mAP}}   & \multirow{2}{*}{0}      & \xmark                          & 68.0   & 34.3  & 33.2    & 53.9     & 51.0       & 70.3 & 9.4   & 71.1       & 39.0       & 69.2   & 81.1   & 24.7 & 46.2    & 87.6 & 73.6 & 54.2                  \\
                         &                         & \cmark                          & 68.9   & 34.7  & 33.8    & 54.5     & 52.6       & 70.5 & 9.3   & 71.7       & 37.7       & 66.7   & 81.7   & 24.3 & 46.6    & 88.9 & 73.3 & 54.3                  \\\cline{2-19}
                         & \multirow{2}{*}{1}      & \xmark                          & 69.9   & 60.3  & 31.4    & 52.2     & 56.3       & 62.8 & 8.8   & 67.7       & 36.4       & 64.7   & 81.6   & 23.4 & 45.9    & 89.2 & 72.1 & 54.8                  \\
                         &                         & \cmark                          & 69.7   & 59.4  & 30.2    & 52.9     & 59.5       & 65.5 & 9.2   & 69.4       & 38.0       & 66.4   & 80.8   & 23.3 & 47.1    & 90.9 & 72.4 & 55.6                  \\\cline{2-19}
                         & \multirow{2}{*}{2}      & \xmark                          & 69.2   & 53.8  & 31.7    & 54.1     & 55.6       & 66.2 & 8.9   & 69.4       & 38.4       & 66.3   & 81.7   & 23.2 & 45.1    & 90.1 & 71.9 & 55.0                  \\
                         &                         & \cmark                          & 69.3   & 54.5  & 32.3    & 54.3     & 56.9       & 66.7 & 9.7   & 69.9       & 38.7       & 66.0   & 81.5   & 24.2 & 48.7    & 90.9 & 72.1 & 55.7                  \\\hline
\multirow{6}{*}{\rotatebox{90}{P-AUC}}   & \multirow{2}{*}{0}      & \xmark                          & 96.5   & 92.7  & 97.4    & 98.1     & 90.7       & 98.2 & 93.1  & 99.0       & 85.7       & 99.0   & 99.5   & 97.9 & 99.3    & 98.7 & 97.7 & 96.2                  \\
                         &                         & \cmark                          & 96.8   & 93.2  & 97.5    & 98.2     & 91.5       & 98.2 & 93.6  & 99.0       & 85.3       & 98.9   & 99.5   & 97.9 & 99.3    & 98.8 & 97.7 & 96.4                  \\\cline{2-19}
                         & \multirow{2}{*}{1}      & \xmark                          & 96.6   & 95.5  & 97.2    & 97.8     & 93.1       & 97.8 & 92.9  & 98.9       & 85.7       & 98.8   & 99.5   & 97.8 & 99.3    & 98.8 & 97.5 & 96.5                  \\
                         &                         & \cmark                          & 96.6   & 95.3  & 97.2    & 97.9     & 93.8       & 97.8 & 93.5  & 98.9       & 86.5       & 98.9   & 99.5   & 98.0 & 99.3    & 99.0 & 97.7 & 96.7                  \\\cline{2-19}
                         & \multirow{2}{*}{2}      & \xmark                          & 96.6   & 95.2  & 97.3    & 98.0     & 92.8       & 98.0 & 93.4  & 98.9       & 86.9       & 98.9   & 99.5   & 97.9 & 99.3    & 98.9 & 97.6 & 96.6                  \\
                         &                         & \cmark                          & 96.6   & 95.2  & 97.4    & 98.0     & 93.1       & 97.9 & 93.5  & 98.9       & 86.9       & 98.9   & 99.5   & 97.9 & 99.4    & 99.0 & 97.6 & 96.7                 \\
\bottomrule
\end{tabular}}
\end{table}

\subsubsection{Ablation study for Learnable Prompts}
The learnable prompt plays an essential role in transferring the rich semantic knowledge of CLIP to anomaly detection. To verify its effectiveness, we replace it with the fixed prompt and perform experiments under various anomaly region refinement layers. As shown in \cref{tablearnrefine}, regardless of the anomaly region refinement layers, adopting learnable prompts always demonstrates steadily improved performance.
Specifically, the P-mAP increases from 54.8\% with fixed prompts to 55.6\% with learnable prompts when one refinement layer is used, indicating that learnable prompts contribute to capturing more comprehensive representations of anomalies, thus promoting the model to localize anomalies more accurately.

\subsubsection{Analysis on tuning different encoders}
We further study the impact of tuning encoders of CLIP~\cite{CLIP}, as shown in \cref{tabtuneclip}. 
We can observe that fine-tuning the text encoder has a negative impact on performance. For example, when we only fine-tune the text encoder, the P-mAP drops from 55.6\% to 52.9\%. 
Additionally, if we fine-tune the image encoder and text encoder simultaneously, the P-mAP is only 53.0\%. And if we keep the text encoder fixed, the performance increases significantly, where P-mAP is from 53.0\% to 55.5\%. 
Notably, adopting the pre-trained model without fine-tuning exhibits optimal performance.
We believe that the main reason lies in that large-scale pre-training gives CLIP rich prior knowledge. However, fine-tuning CLIP might damage this prior knowledge due to the substantial domain gap between industrial data and natural images.

\subsubsection{The Influence of Data Scale}

\begin{table}[t]
\centering
\caption{The influence of fixing different encoders in CLIP.}
\label{tabtuneclip}
\setlength{\tabcolsep}{8pt}

\begin{tabular}{ccccc}
\toprule
Tune Text Encoder & Tune Image Encoder & I-AUC                                 & P-AUC      & P-mAP                    \\
\midrule
\cmark                  & \cmark                 & 96.3                                 & 96.2                                 & 53.0                                 \\
\xmark                 & \cmark                 & \textbf{97.5}                                & 96.6                                 & 55.5                                 \\
\cmark                  & \xmark                & 96.4                                 & 95.9                                 & 52.9                                  \\
\xmark                 & \xmark          & \textbf{97.5} &\textbf{96.7} & \textbf{55.6} \\
\bottomrule
\end{tabular}
\end{table}

\begin{figure*}[t]
\centering
\includegraphics[width=0.95\textwidth]{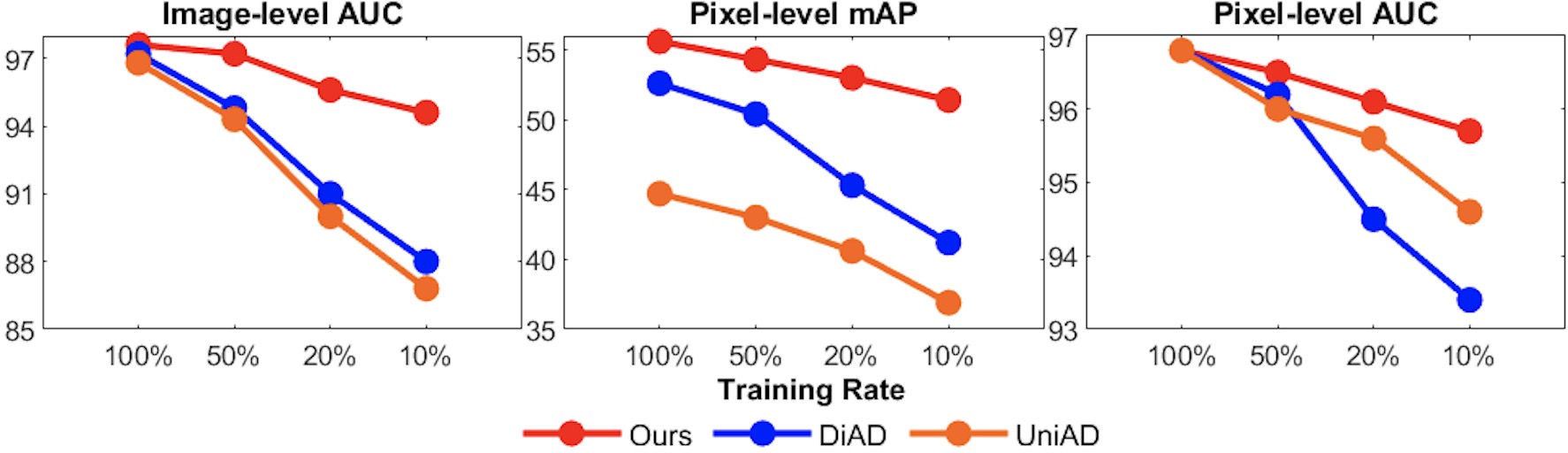}
\caption{Ablation study for the influence of data scale.}
\label{tabdatascale}
\end{figure*}

CLIP has been trained on large-scale image-text datasets, and it can extract high-quality generalizable representations that help alleviate the severe data shortage problem in the industrial anomaly detection field. We further conduct additional experiments to evaluate the effectiveness of our framework, using different data scales. We vary the ratio of training samples from 10\% to 50\% and compare the results with previous arts. As shown in \cref{tabdatascale}, our method consistently demonstrates superior performance across all metrics, surpassing previous arts comprehensively. These impressive results verify the generalization capacity of our method and emphasize its practical value for real-world applications. More details are available in the Supplementary.

\subsubsection{Analysis of computational complexity}
~\cref{tbcomplex} compares the computational complexity between our method and other previous arts. In our method, it can be observed that as the refinement layer increases, the performance gradually increases along with the computational complexity. Specifically, as the layers increase to two, the FPS decreases from 65.0 to 35.0 img/s. However, compared with existing methods, it still maintains a competitive inference speed.
Moreover, our method exhibits significant parameter efficiency, with a model size (1.2M at most) considerably smaller than other previous arts like DiAD~\cite{he2023diad}.

\begin{table}[t]
\centering
\setlength{\tabcolsep}{7pt}
\caption{Computational complexity comparison. \emph{Num} denotes the number of refinement layers. \emph{Params} denotes the trainable parameters.}
\resizebox{0.85\textwidth}{!}{
\label{tbcomplex}
\begin{tabular}{ccccccc}
\toprule
Method                & \emph{Num}        & I-AUC & P-AUC & P-mAP & \emph{Params} (M) & FPS (img/s) \\
\midrule
RD4AD                 & \multirow{4}{*}{/} & 94.6  & 96.1  & 48.6  & 92.2       & 84.1        \\
DRÆM                 &                    & 88.1  & 87.2  & 52.5  & 97.4       & 45.9        \\
UniAD                 &                    & 96.5  & 96.8  & 43.4  & 7.5        & 27.6        \\
DiAD                  &                    & 97.2  & 96.8  & 52.6  & 1331.3     & 0.9         \\\hline
\multirow{3}{*}{Ours} & 0                  & 97.4  & 96.4  & 54.3  & 0.4        & 65.0        \\
                      & 1                  & 97.5  & 96.7  & 55.6  & 0.8        & 46.2        \\
                      & 2                  & 97.5  & 96.7  & 55.7  & 1.2        & 35.0       \\
\bottomrule
\end{tabular}}
\end{table}

\section{Limitation} 
Despite our method demonstrating superiority by using learnable text prompts and a coarse-to-fine strategy, it encounters limitations within complex and variable scenarios, indicating the need for improving robustness to handle complex and varied scenarios.

\section{Conclusion}
In this paper, we have proposed a simple yet effective anomaly detection framework, named CLIP-ADA. 
Directly applying CLIP to anomaly detection field poses significant challenges due to the domain gap between the pre-training data and industrial images, alongside the requirement for more fine-grained representations.
To address this challenge, we construct learnable text prompts and adaptively associate them with normal regions through self-supervised learning.
Additionally, we also suggest a coarse-to-fine strategy to further refine anomaly regions.
Experimental results show that CLIP-ADA obtains leading anomaly detection and localization performance.

\clearpage  %

\bibliographystyle{splncs04}
\bibliography{main}
\end{document}